\documentclass[a4paper,11pt]{article}

\usepackage[english]{babel}
\usepackage[utf8]{inputenc}
\usepackage[T1]{fontenc}
\usepackage{lmodern}
\usepackage{microtype}

\usepackage[a4paper,top=3cm,bottom=2.5cm,left=3cm,right=3cm,marginparwidth=1.75cm]{geometry}
\usepackage[parfill]{parskip} 
\usepackage{setspace}
\usepackage{amsmath,mathtools}
\usepackage{graphicx}
\usepackage{booktabs}
\usepackage{tabularx}
\usepackage{caption}
\usepackage{subcaption}
\usepackage{float}
\usepackage{multirow}
\usepackage{threeparttable} 
\usepackage{calc}

\usepackage{siunitx}
\usepackage[dvipsnames,table]{xcolor}
\definecolor{sunflower}{HTML}{FFC107} 
\definecolor{graphite}{HTML}{222222}  
\definecolor{slate}{HTML}{556172}     

\usepackage{hyperref}
\hypersetup{
  colorlinks=true,
  linkcolor=blue,     
  urlcolor=blue,      
  citecolor=blue      
}
\usepackage{cite}
\usepackage{natbib}

\usepackage{titlesec}
\titleformat{\section}{\large\bfseries\color{graphite}}{\thesection}{0.75em}{}
\titleformat{\subsection}{\normalsize\bfseries\color{graphite}}{\thesubsection}{0.75em}{}
\titleformat{\subsubsection}{\normalsize\bfseries\color{graphite}}{\thesubsubsection}{0.75em}{}
\titlespacing*{\section}{0pt}{0.6em}{0.4em}
\titlespacing*{\subsection}{0pt}{0.5em}{0.3em}
\titlespacing*{\subsubsection}{0pt}{0.4em}{0.2em}

\usepackage{fancyhdr}
\title{\vspace{-2cm}Sunflower: A New Approach To Expanding Coverage of African Languages in Large Language Models}

\author{Benjamin Akera, Evelyn Nafula Ouma, Gilbert Yiga, Patrick Walukagga, \\
Phionah Natukunda, Trevor Saaka, Solomon Nsumba, Lilian Teddy Nabukeera, \\ Joel Muhanguzi, Imran Sekalala,  Nimpamya Janat Namara, \\ Engineer Bainomugisha, Ernest Mwebaze, John Quinn \\ \\Sunbird AI, Uganda \\\url{info@sunbird.ai}}

\begin{document}




\maketitle

\begin{abstract}
There are more than 2000 living languages in Africa, most of which have been bypassed by advances in language technology. Current leading LLMs exhibit strong performance on a number of the most common languages (e.g. Swahili or Yoruba), but prioritise support for the languages with the most speakers first, resulting in piecemeal ability across disparate languages. We contend that a regionally focussed approach is more efficient, and present a case study for Uganda, a country with high linguistic diversity. We describe the development of Sunflower 14B and 32B, a pair of models based on Qwen 3 with state of the art comprehension in the majority of all Ugandan languages. These models are open source and can be used to reduce language barriers in a number of important practical applications.
\end{abstract}

\begin{figure}[b!]
    \centering
    \includegraphics[width=1\linewidth]{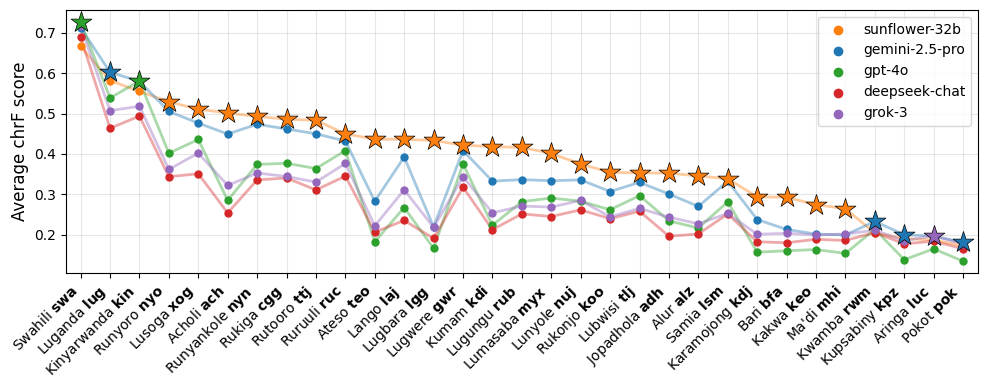}
    \caption{Comparison of machine translation performance (mean chrF over xx$\rightarrow$eng and eng$\rightarrow$xx). Sunflower 32B has the highest accuracy in 24 out of 31 Ugandan languages.}
    \label{fig:comparison_chrf}
\end{figure}

\section{Introduction}

When LLMs are developed for a global audience by centralised teams, languages are prioritized based on such factors as speaker numbers, data availability, and potential revenue. While these are natural considerations, particularly in a commercial context, this results in patchy support for just a few of the world's several thousand languages. Even the best-resourced, flagship models have official support for only a handful of languages: ChatGPT, for example, officially supports 59 languages at the time of writing \cite{chatgptlanguages2025} (though has unofficial capability for more). It remains unclear how long it will take for this centralized development model to yield LLMs capable of communicating effectively in the majority of the world’s languages.

In this paper, we present an alternative approach to LLM multilinguality: rather than including several disparate languages in one model, to focus on a single region with high linguistic diversity and to include all the languages spoken there. Our case study is the 40 or so languages spoken in Uganda, and we describe the development of Sunflower 14B and 32B, a pair of instruction-tuned models based on Qwen 3 \cite{yang2025qwen3}, which are able to carry out translation, question-answering and a wide range of creative tasks. The motivations for this approach are:

\begin{itemize}
    \item By focusing on languages within a single region, we leverage shared linguistic structures, cultural concepts, and contextual knowledge. Many languages in our study do not have significant data resources on their own, but are related to other languages spoken in neighbouring regions, so that there can be effective learning transfer.
    \item A team who are physically located in the region being studied is in a strong position to form networks of data contributors, advisors and evaluators. It is then possible to use datasets which come in unconventional forms, e.g. printed matter not available online, or radio stations contributing audio recordings that can be transcribed.
    \item A team local to the region, who speak the languages being modelled, are able to assess failure modes with greater speed and nuance than a remote team relying on third party contractors for data labelling and evaluation.
\end{itemize}

Our main focus in this work was on machine translation performance, as a quantifiable measure of cross-lingual understanding, for which we achieve state-of-the-art performance in 24 out of 31 evaluation languages (see Figure \ref{fig:comparison_chrf} and Section~\ref{sec:eval_and_results}). We also carry out instruction tuning and demonstrate the models' ability to respond to a wide variety of multilingual task prompts.

In the remainder of this paper: we discuss related work in \S\ref{sec:related_work}, then describe our data sources in \S\ref{sec:data_sources}, training and evaluation datasets in \S\ref{sec:datasets}, and model training methodology in \S\ref{sec:model_training}. Details of our evaluation are in \S\ref{sec:eval_and_results} comparing Sunflower against alternatives, followed by discussion of findings and future directions in \S\ref{sec:discussion}.

The Sunflower models are available open source (\url{https://hf.co/Sunbird}), such that they can be freely used for a variety of practical purposes to overcome language barriers in Uganda, and to help accelerate research in African language technologies.

\begin{table}[h!]
\small
\centering
\caption{Sample responses to a question posed in the Acholi language: Sunflower 14B responds informatively whereas ChatGPT fails to correctly interpret the question.}
\begin{tabular}{>{\raggedright\arraybackslash}p{0.45\textwidth}>{\raggedright\arraybackslash}p{0.45\textwidth}}
\toprule
\multicolumn{2}{c}{Prompt: Atwero penyo nining pi karatac me wot i lobo Uganda?} \\
\multicolumn{2}{c}{(\emph{How do I apply for a passport in Uganda?})} \\
\midrule
Sunflower 14B & ChatGPT \\
\midrule
Me penyo pi karatac me wot i Uganda, itwero timo jami magi:
\vspace{0.5em}

1. Cit i opic pa lacungu pi lobo Uganda: Itwero cito bot lacungu pi lobo Uganda i lobo ma itye iye, i opic me Uganda, onyo i opic ma cung pi lobo Uganda. Gamente mapat-pat bene twero miyo kony.
\vspace{0.5em}

2. Coo ne i kabedo me intanet: Itwero coyo karatac me wot i Uganda kun itiyo ki kabedo me intanet me Uganda.
\vspace{0.5em}

3. Coo ne i opic pa lacungu pi lobo Uganda: Ka itye i lobo ma pe jenge i kom gamente, itwero coyo karatac me wot i Uganda kun itiyo ki opic pa lacungu pi lobo Uganda matye cok kwedi.

& Mai i kemi ka ma lim-tok, ma tye ki “Atwero penyo nining pi karatac me wot i lobo Uganda” pe! Kadi ngano pe, an isi kwena kel mawoto.
\vspace{0.5em}

“Penyo” me Acholi pe, ngano tye ki “how many / what number”. “Atwero penyo” ngambe tye ki “how far” / “what distance”. “Karatac” tye ki “kilometres”. “Wot” tye ki “from”. “Lobo” tye ki “city / town / place”.
\vspace{0.5em}

Ka acaki ngat acel acel, kit ma pe ce ki tye: “How many kilometres is it from Lobo, Uganda to some place called Atwero?” To answer properly, I need to know which Atwero you mean (there may be more than one) and which “Lobo” (there’s perhaps a place called Lobo).
\vspace{0.5em}

If you mean Lobo in Moroto County, Alebtong District, Northern Uganda, there is “Lobo Oderoryek” village near Abako. [...] \\ 

\vspace{.7em} & \\

(\emph{To apply for a Ugandan passport, you can do the following:
1. Go to the office of a government representative: [...]
}) & 
(\emph{Reponse mentions the correct language Acholi, but is not easily interpretable, and misunderstands the question.})
\\
\bottomrule
\end{tabular}
\label{tab:uganda_responses}
\end{table}

\section {Related Work}
\label{sec:related_work}

Our work builds upon three key research areas: datasets for African languages, multilingual language model development, and specialized techniques for low-resource language processing. While significant progress has occurred in each area, gaps remain in comprehensive regional language modeling and practical deployment for Ugandan languages.

\subsection{African language datasets and resources}

Recent years have witnessed substantial efforts to create datasets for African languages. The Masakhane initiative \cite{orife2020masakhane} has coordinated translation dataset creation across multiple African languages, with projects like MAFAND-MT \cite{adelani-etal-2022-thousand} providing parallel corpora for African news translation while comprehensive surveys \cite{adebara2022towards} document the current state of African NLP resources and identify key gaps. Large-scale multilingual datasets like MADLAD-400 \cite{kudugunta2023madlad} include text from hundreds of languages, and FLORES-200 \cite{costa2022no} offers evaluation benchmarks for machine translation across diverse language pairs. 
However, these resources face three key limitations for Ugandan language applications. First, coverage remains sparse; While MADLAD-400 includes some Ugandan languages, deep neural networks require large amounts of training data, and the shortage of large-scale data in low-resource languages makes their processing challenging \cite{hedderich2020survey}. Second, existing datasets emphasize translation pairs over comprehensive language modeling data \cite{adelani-etal-2022-thousand}. Third, most resources lack the cultural context and domain-specific terminology crucial for practical applications in education and healthcare \cite{nekoto2020participatory}. The SALT corpus \cite{akera2022machine} represents the most comprehensive effort targeting Ugandan languages specifically, providing parallel text across multiple domains. Building on SALT's foundation, we extend coverage to 40+ languages and introduce instruction-following formats absent from existing African language resources.

\subsection{Multilingual language models and evaluation}

Multilingual language models have evolved from early approaches like mBERT \cite{devlin2019bert} to large-scale models including BLOOM \cite{workshop2022bloom}, mT5 \cite{xue2020mt5}, No Language Left Behind (NLLB) \cite{costa2022no}, and recent multilingual variants of Llama \cite{dubey2024llama}, Qwen \cite{bai2023qwen}, and Gemma \cite{team2024gemma}. While these models demonstrate impressive zero-shot transfer capabilities across languages, they exhibit systematic performance degradation on low-resource languages. Pre-trained multilingual language models often fail to provide adequate support for languages with small available monolingual corpora, leading to poor performance  \cite{hangya2022improving}. This limitation stems from fundamental challenges in multilingual training, including imbalanced training corpus distribution that creates bias toward common languages  and parameter sharing conflicts that cause negative language interference \cite{li2024quantifying}.

Evaluation frameworks like XTREME \cite{hu2020xtreme}, XTREME-R \cite{ruder2021xtreme}, COMET \cite{rei2020comet}, AfriMTE and AfriCOMET \cite{wang2023afrimte}, and GEM \cite{gehrmann2021gem} provide standardized benchmarks for multilingual model assessment. However, these frameworks face significant limitations for African language applications. XTREME covers 40 typologically diverse languages but includes only Swahili and Yoruba among African languages, while XTREME-R extends coverage to 50 languages with similarly limited African representation. Moreover, these benchmarks emphasize standard NLP tasks like sentence classification and structured prediction rather than the culturally-specific applications essential for local communities, such as healthcare consultations or traditional knowledge preservation.

Recent work has explored continued pretraining for improved low-resource language performance \cite{joshi2020state},\cite{owodunni2025continually},\cite{li2025rethinking} and adapter-based approaches for efficient multilingual fine-tuning \cite{khade2024challenges}, \cite{hu2022lora}, \cite{liang2025towards}. Continued pretraining approaches demonstrate that multilingual LLMs can be effectively adapted to low-resource languages through strategic data mixing and synthetic corpus augmentation, while parameter-efficient fine-tuning methods like LoRA show promise for adapting multilingual models to specific languages while maintaining computational efficiency. However, these methods typically target individual languages in isolation rather than exploiting regional linguistic similarities and shared cultural contexts as demonstrated in our approach with Ugandan languages.

\subsection{Regionally-focused multilingual models}

Recent initiatives have begun exploring regional specialization as an alternative to global multilingual coverage. SEA-LION \citep{ng2025sea}, developed by AI Singapore, targets 11 Southeast Asian languages including Indonesian, Thai, Vietnamese, Filipino, Burmese, Malay, Tamil, Lao, and Khmer. The project employs continued pretraining on regionally-specific corpora, expanding vocabulary to improve tokenization efficiency for non-Latin scripts. SEA-LION v4 (27B parameters) achieves competitive performance with much larger models while maintaining computational efficiency for deployment in resource-constrained environments.

Parallel efforts include SeaLLM \citep{nguyen2023seallms}, developed by Alibaba DAMO Academy, which builds on Llama 2 with continued pretraining on 8 Southeast Asian languages. SeaLLM demonstrates particular strength in processing non-Latin scripts, achieving up to 9$\times$ compression ratios compared to models optimized for English. Sailor \citep{dou2024sailor}, developed collaboratively by SEA AI Lab and Singapore University of Technology and Design, focuses on Indonesian, Thai, Vietnamese, Malay, and Lao, with model sizes ranging from 0.5B to 20B parameters built through continued pretraining of Qwen base models.

These Southeast Asian initiatives share several characteristics: (1) focus on a coherent linguistic region with shared cultural contexts, (2) continued pretraining from strong multilingual base models rather than training from scratch, and (3) emphasis on practical deployment considerations including model size optimization and efficient tokenization. Their geographical scope is wider than what we consider in this work, spanning multiple nations with diverse language families.

For African languages, InkubaLM \citep{tonja2024inkubalm} demonstrates efficiency through a compact 0.4B parameter model covering Swahili, Yoruba, Hausa, isiZulu, and isiXhosa. While achieving competitive performance on sentiment analysis and translation tasks despite minimal parameters and training data, InkubaLM adopts a continental rather than regional scope, covering languages from disparate geographic areas across Africa without exploiting potential linguistic similarities within smaller regions.

Our work extends the regional specialization paradigm in two critical dimensions. First, we focus on languages within a single country (Uganda) rather than a broader geographic region, enabling deeper exploitation of shared linguistic structures and cultural knowledge. Second, we address languages with significantly lower digital resources than those targeted by Southeast Asian models, requiring novel data collection strategies beyond web scraping. This approach demonstrates that extreme regional specialization—targeting all languages within a single nation—can achieve state-of-the-art performance while serving communities typically overlooked by both global and broader regional initiatives.

\subsection{Instruction tuning and task-specific datasets}

The development of instruction-following language models has been driven by carefully curated datasets that enable models to understand and execute natural language instructions across diverse tasks. Foundational instruction datasets include FLAN \cite{longpre2023flan}, which demonstrated the effectiveness of instruction tuning by training models on over 60 NLP datasets formatted with natural language prompts, and Alpaca \cite{alpaca}, which provided 52,000 instruction-response pairs generated through self-instruction techniques. The Databricks Dolly 15k dataset \cite{DatabricksBlog2023DollyV2} further advanced the field by offering 15,000 human-generated instruction examples specifically designed for commercial use, while other collections have contributed additional task diversity and scale.
Large-scale pretraining datasets like The Pile \cite{pile} and the Colossal Clean Crawled Corpus (C4) \cite{raffel2020exploring} provide the foundation for language model training, but these resources focus overwhelmingly on English and other high-resource languages. The Pile, comprising 825 GiB of diverse English text from 22 sources, and C4, derived from Common Crawl web data, have enabled remarkable progress in English language modeling. However, recent efforts to create multilingual instruction datasets remain limited in scope for African languages, with existing collections providing minimal coverage and task diversity compared to their English-focused counterparts.

\subsection{Low-Resource NLP techniques} 

Specialized techniques for low-resource language processing have emerged to address the challenges of limited training data and evaluation resources. Data augmentation approaches, particularly back-translation \cite{sennrich2015improving}, have proven effective for generating synthetic parallel data, while cross-lingual transfer learning methods leverage shared representations across linguistically related languages. Parameter-efficient fine-tuning techniques, including LoRA and adapter-based methods \cite{hu2022lora}, \cite{pfeiffer2020adapterhub}, enable resource-efficient adaptation of large multilingual models to specific languages or domains. 
Recent research has explored optimal data collection strategies for low-resource languages, including continued pretraining approaches \cite{joshi2024adapting} and synthetic data generation techniques. Existing techniques typically target individual languages in isolation, which do not exploit regional linguistic similarities and shared cultural knowledge that could benefit grouped language families. This limitation becomes particularly pronounced for African languages, where related languages within geographical regions often share structural features, borrowed vocabulary, and cultural concepts.

\subsection{Contributions of this work}

Our work represents a change of emphasis from massive, centralised multilingual models toward specialized, regionally-coherent language modeling. We demonstrate that with relatively small models, it is possible to outperform existing systems for the specific task of machine translation, and that instruction tuning can be carried out to give the model the ability to respond to a wide range of instruction types. This is due to the following factors:

\begin{enumerate}
    \item \textbf{Regional linguistic coherence:} Rather than treating each language as an isolated entity, we exploit the  linguistic and contextual coherence inherent across Uganda’s 40+ languages. Languages have similar structural features (such as agglutinative morphology prevalent across many language groups), regional phonetic overlaps, and, crucially, common cultural concepts and entities specific to the region that often lack direct English equivalents. By focusing on this regional grouping which includes languages spanning the Bantu, Nilotic, and Central Sudanic families, we enable more efficient parameter sharing and robust cross-lingual transfer compared to models that must simultaneously handle globally disparate language groups. 

    \item \textbf{Multi-source data acquisition beyond web dependency:} We use a variety of methods for collecting data sources in many languages, including digitization of printed materials through OCR, transcription of radio programming, and community-sourced cultural documents. The focus on one particular country helps with this.

    \item \textbf{Culturally-grounded evaluation framework:} We evaluate our models on examples which are representative of day-to-day modes of communication in different scenarios (Section \ref{sec:evaluation_data}), and thus match the context in which they will be deployed. We also emphasise community feedback. 
\end{enumerate}

We demonstrate that specialized, regionally-focused models can achieve superior performance on culturally-relevant tasks while requiring significantly fewer computational resources than their globally-oriented counterparts. In principle, our work could also be replicated in any country or region with high linguistic diversity. The resulting open source Sunflower models are useful either for fine-tuning in specific applications, or for extending to other languages in the region. 

\section{Data sources}
\label{sec:data_sources} 

We curated training data in three primary categories: existing digital corpora, digitized physical media through OCR and audio transcription, and specialized linguistic resources including dictionaries and community documents. We found that significant data on Ugandan languages had to be collected from printed materials, audio recordings, and cultural archives, not being easily accessible via the web. This section describes our data acquisition methodology; subsequent sections detail the processing of this raw data into pretraining and instruction datasets.

We began by collecting existing digital corpora to establish a foundation for the dataset. Large-scale web text from MADLAD-400 \cite{kudugunta2023madlad}, Ugandan news articles via Common Crawl, and established multilingual datasets with parallel text across languages including FLORES-200 \cite{nllb2022}, MT560 \cite{gowda2021many}, Google SMOL \cite{caswell2025smol}, TICO-19 \cite{shardlow2022simple}, MAFAND-MT \cite{adelani-etal-2022-thousand} and our own SALT dataset \cite{akera2022machine}. For this work we extended the SALT dataset, which contains translations for 25,000 sentences, to add professional translations in Rutooro, Runyoro and Swahili, adding to our previous version with translations in Luganda, Acholi, Ateso, Runyankore and Lugbara. Bible texts were also included, where we found translations for 28 languages that could be aligned verse by verse (although with a caveat that some translations appeared to use archaic forms of the languages which did not correspond to modern usage). We also incorporated the Makerere MT corpus \cite{mukiibi_makerere_2021}, Mozilla Common Voice \cite{ardila2019common}, and Uganda-specific Q\&A examples in the Ugandan Cultural Context Benchmark (UCCB) Suite \cite{uccb_2025}.

A substantial portion of linguistic data for Ugandan languages exists only in physical/printed or audio formats, inaccessible through conventional web-based collection methods. We therefore conducted extensive digitization of printed materials including out-of-print books, educational textbooks, and language guides through optical character recognition (OCR). In partnership with the local radio station Simba FM, we transcribed over 500 hours of audio recordings of talk shows across several topics, using a version of Whisper Large v3 which we fine-tuned to transcribe speech in ten Ugandan languages.

We also sourced and digitized dictionaries, providing structured data on lexical semantics and morphology, and collected proverbs, traditional folklore, and culturally-specific texts from community archives and publications. 

\section{Training and evaluation datasets}
\label{sec:datasets}

The above collection and curation process yielded two complementary datasets for different training stages: a large-scale pretraining corpus and a structured instruction-tuning collection. The pretraining corpus enables the model to adapt to Ugandan languages' vocabulary, syntax, and cultural knowledge, while the instruction dataset aligns the model with practical tasks. This two-stage approach leverages our diverse source materials, from digitized books to radio transcripts, first for language adaptation, then for task-specific alignment.

\begin{table}[t!]
\footnotesize
\centering
\caption{Extracts from factuality preferences training dataset.}
\label{tab:african_qa}
\begin{tabular}{>{\raggedright\arraybackslash}p{4cm}>{\raggedright\arraybackslash}p{5cm}>{\raggedright\arraybackslash}p{5cm}}
\toprule
\textbf{Prompt} & \textbf{Preferred response} & \textbf{Dispreferred response} \\
\midrule
What is the title of the king of the Kingdom of Buganda? & The title of the king of the Kingdom of Buganda is Kabaka. & The title of the king of the Kingdom of Buganda is Mwami. \\
\addlinespace
Who became Burundi's first Hutu president in 1993? & Melchior Ndadaye became Burundi's first Hutu president after winning the June 1993 election; he was assassinated in October 1993. & Adrien Ruziki became Burundi's first Hutu president after winning the July 1993 election; he was assassinated in December 1993. \\
\addlinespace
How many harvests per year can Pajok's climate allow? & The climate in Pajok can support two and sometimes three harvests each year. & The climate in Pajok only allows for a single harvest each year. \\
\bottomrule
\end{tabular}
\label{tab:factuality_dataset}
\end{table}

\begin{table}[t]
\footnotesize
\centering
\caption{Extracts from glitching preferences training dataset.}
\label{tab:african_qa}
\begin{tabular}{>{\raggedright\arraybackslash}p{4cm}>{\raggedright\arraybackslash}p{5cm}>{\raggedright\arraybackslash}p{5cm}}
\toprule
\textbf{Prompt} & \textbf{Preferred response} & \textbf{Dispreferred response} \\
\midrule
In Rukonjo, please draft a compelling cover letter for a graphic design position. & **Omuhya w'ebisisani - Ebaruha y'eriyimba** Ab'ekithunga kya [Erina ly'ekithunga], baghe, Ngahandika eriyimba eriyisunza lyaghe omwa kitsweka ky'omuhya w'ebisisani ebyakangahirirwa okwa [Erina ly'ekithunga]. Nga [...] & **Namalira y'omubiiri w'omubiiri w'omubiiri w'omubiiri w'omubiiri w'omubiiri w'omubiiri w'omubiiri w'omubiiri w'omubiiri w'omubiiri w'omubiiri w'omubiiri w'omubiiri w'omubiiri w'omubiiri w'omubiiri w'omubiiri w'omubiiri   [...] \\
\addlinespace
Mpa olukalala lw'ebirowoozo eby'entegeka y'okukola okulongosa awaka okw'okumala wiikendi. & Gano niigo amagondi g'ebintu eby'okukola eby'okulongosa eka ebyo ebyotakola okunkomerero y'esabbiiti: 1. **Okuteekawo emikeeka gy'omu kiyungu**: Woba oli n'ekiyungu ekya ngulu eky'okuteekamu, kisobola okubba pulojekiti y'okunkomerero  [...] & Nga weekwata ennyumba yo, weetaaga okusalawo mu ngeri ey'amagezi ku ngeri y'okugikuuma nga terina buzibu mu biseera eby'omu maaso. Wammanga wammanga wammanga wammanga wammanga wammanga wammanga wammanga  wammanga [...] \\
\bottomrule
\end{tabular}
\label{tab:glitching_dataset}
\end{table}

\subsection{Pretraining text}
\label{sec:pretraining_corpus}

The pretraining corpus adapts the base language model to Ugandan languages' vocabulary, syntax, and cultural/factual knowledge. We combined cleaned text from all collected sources, including digitized books, web articles, and transcribed audio.

We applied normalization and cleaning to standardize the corpus while preserving critical linguistic information, including normalizing whitespace, removing control characters, and eliminating digitization artifacts such as page headers and OCR errors. We standardized character encodings while preserving language-specific diacritics and special characters essential for accurate representation of Ugandan languages. This was necessary as we found that when incorporating pdfs from diverse sources, there was a tendency for unicode characters to be erroneously represented. OCR-based data needed particular scrutiny due to challenges such as complex page layouts or misrecognition of text with unusual character sets.

The language distribution of this corpus is inevitably imbalanced, as some languages have much greater numbers of speakers. During prototyping, we found that adding back-translated text helped to improve performance for some of these less-represented languages. To do this, we trained a separate machine translation model based on NLLB 3.3B on all parallel text. Then we translated English sentences to selected languages with less representation, and added these to the pretraining text.

We also include examples of instructions in conversational format within this pretraining text, to prevent catastrophic forgetting of the original instruction tuning.

In total, around 1B characters of pretraining text were collected. To maximise training efficiency, we used extra steps to avoid duplication of any text and to compress it by removing as much extraneous detail as possible. This reduced the pretraining text to around 0.75B characters.

\subsection{Conversational instruction examples} \label{sec:instruction_tuning}

The instruction-tuning collection aligns the pretrained model with practical tasks by transforming our source materials into supervised examples of user prompts and assistant responses. We structured this collection around five task categories that leverage our diverse data sources

\begin{enumerate}
    \item \textbf{Translation:} Bidirectional translation between English and Ugandan languages, covering multiple domains from general text to biblical verses. To improve robustness, some tasks simulate noisy inputs, such as those from an Automatic Speech Recognition (ASR) system.

    \item \textbf{Question Answering:} Contextual question answering derived from educational language guides and other informative documents.

    \item \textbf{Summarization and Correction:} Abstractive summarization and text correction tasks, specifically designed to handle and denoise imperfect inputs, such as raw ASR transcripts.

    \item \textbf{Creative Tasks:} A variety of generative tasks, such as examples of writing poems or articles was included.

    \item \textbf{Cultural Explanation:} Tasks requiring the model to explain cultural concepts, proverbs, and region-specific knowledge, thereby grounding its capabilities in the local context.    
\end{enumerate}

\subsection{Evaluation data}
\label{sec:evaluation_data}

We curated a new dataset with translations across 31 Ugandan languages, in collaboration with Makerere University Institute of Languages. These were all the languages for which we were able to identify a professional translator. The dataset was created by first generating English mini-documents/dialogues composed of 5 sentences for each of 20 different categories. The categories are listed in Table \ref{tab:evaluation_categories}, with an example of LLM-generated sentences in one category shown in Table \ref{tab:evaluation_sentences}. Each sentence was then translated into all of the 31 languages in our evaluation set. In total we therefore have 3100 translated sentences, which give 6200 evaluation points if we only consider translation directions to English or from English.

This scheme was chosen because (1) it allows evaluation at either the sentence level or the paragraph/longer-form level when they are concatenated, and (2) we wish to cover many different styles of communication -- formal, informal, dialogue, news, fiction, and so on.

\begin{table}[t]
\small
\centering
\caption{Categories of evaluation text.}
\begin{tabular}{p{0.05\textwidth}p{0.4\textwidth}p{0.05\textwidth}p{0.4\textwidth}}
1. & Banking transaction dialogue & 11. & Health advice, e.g. obstetrics \\
2. & Instructional/education material & 12. & Official announcement \\
3. & Conversation with typical greetings & 13. & Speech given by official \\
4. & Fiction/story & 14. & Farming advice \\
5. & Medical scenario, doctor/patient & 15. & Response to opinion poll \\
6. & News story & 16. & Public transport scenario \\
7. & Wikipedia style information & 17. & Family \\
8. & Everyday shopping/transactions & 18. & Nutrition \\
9. & Emergency response & 19. & Food security advice \\
10. & Practical guide for mechanic or builder & 20. & Local council dialogue \\
\end{tabular}
\label{tab:evaluation_categories}
\end{table}

\begin{table}[t]
\footnotesize
\centering
\caption{Generated English text for scenario 10 (practical guide for mechanic or builder).}
\begin{tabular}{p{1.0\textwidth}}
First, make sure the foundation trench is at least two feet deep and well compacted before laying any bricks. \\
Mix the cement, sand, and aggregate in a 1:2:4 ratio for strong concrete, especially for the pillars. \\
Use a spirit level to keep the walls straight and check alignment after every few layers of bricks. \\
Keep the site clean and water the curing concrete regularly for at least seven days to prevent cracks. \\
Finally, make sure all electrical and plumbing points are marked clearly before plastering the walls. \\
\end{tabular}
\label{tab:evaluation_sentences}
\end{table}

\section{Model training}
\label{sec:model_training}

The training process for the Sunflower models has three phases: (1) continued pretraining to add knowledge of Ugandan languages and relevant factual knowledge, then (2) supervised fine-tuning to train it to follow instructions, with particular emphasis on translation, and finally (3) post-training adjustment with reinforcement learning to improve model responses to conversational prompts.

\subsection{Continued pretraining}
\label{sec:pretraining}

Continued pretraining of the base Qwen 3 models are used with the data as detailed in Section \ref{sec:pretraining_corpus}. The training objective is next token prediction on this wide range of text data, which includes parallel text in multiple languages, books and documents, transcripts from audio, and various other sources. This instills the model's basic capability in comprehending and generating text in our languages of interest, as well as injecting knowledge about relevant culture, history, events, and concepts.

Continued pretraining of the 32B model took around 8 hours on 8 x H200 GPUs, and the 14B model around 6 hours on 4 x H200 GPUs. In both cases, DeepSpeed ZeRO-3 was used to fit the full models into VRAM. FSDP was also tried but resulted in higher validation loss. The effectiveness of continued pretraining was highly sensitive to learning rate and batch size. We use a batch size of 32,768 tokens, arranged in memory as 64 sequences of 512 tokens, using sample packing for efficiency, a warmup ratio of 0.01 and peak learning rate of 1e-4, for a single epoch.

\subsection{Supervised fine-tuning}
\label{sec:fine-tuning}

Supervised fine tuning was carried out on a mixture of translation instructions (N=(2347), which was our primary training goal, and general responses to conversational queries (N=726), as described above. For this we use a LoRA adapter with rank 16, which has the dual benefit of reducing the VRAM requirement while training and acting as a form of regularisation. We train for one epoch with batch size 16 and learning rate 1e-5.

We train with loss only on the response part of each example, i.e. we do not train the model on the text of the instruction, or the tokns in the chat template. Without this, the models exhibit a tendency to repeat the user's instruction. We also constructed examples in which multiple instructions were concatenated into a single conversation, to address another issue with prototype models in which a user would switch topic but the model would still refer to previous questions.

\subsection{Reinforcement learning with Direct Preference Optimisation}
\label{sec:reinforcement_learning}

Supervised fine-tuning gave strong results for machine translation, but the resulting models exhibited problems as conversational assistants. Particular issues were glitching (especially a tendency for the model to get stuck in infinite loops) and hallucinations.

We applied Direct Preference Optimisation (DPO) to mitigate these problems. DPO takes training examples structured as sample prompts with both preferred (or `chosen') responses and dispreferred (or `rejected') responses. Gradient descent is then used to align the model to these preferences. We used a variant of DPO loss, Iterative Reasoning Preference Optimization (Pang et al, 2024), setting $\alpha_{RPO}$=1.0. This essentially mixes in a supervised fine-tuning loss to add positive weight to the preferred examples, which we found stabilised the training and produced subjectively better results than standard DPO.

\section{Evaluation}
\label{sec:eval_and_results}

Sunflower's performance is evaluated on translation tasks across 31 Ugandan languages, comparing against leading proprietary and open-source models. This section describes our methodology, presents results, and discusses the model's broader capabilities beyond machine translation.

\begin{table}[t]
\footnotesize
\centering
\caption{Sample responses from Sunflower 14B to freeform prompts. Translations shown in italics, and outputs abbreviated for clarity.}
\begin{tabular}{l>{\raggedright\arraybackslash}p{0.8\textwidth}}
\toprule
\bf Prompt: & Explain how photosynthesis works in Luganda \\
\bf Response: & Photosynthesis kwe kusobozesa ebimera okukola emmere yaabyo nga byeyambisa ekitangaala, amazzi, n'omukka oguyitibwa carbon dioxide. Ebikoola by'ebimera birina obutuli obuyitibwa stomata mwe biyingiriza omukka ogwo. [...] \vspace{0.1cm} \\
& (\emph{Photosynthesis is the process by which plants make their own food using light, water, and carbon dioxide. The leaves of plants have tiny holes called stomata that allow them to take in air} [...]) \\
\midrule
\bf Prompt: & Nnyonnyola x kye ki singa x$^2$ + 2 = 6? \\ 
& \emph{(Explain what x is if x$^2$ + 2 = 6?)} \vspace{0.1cm} \\
\bf Response: & Okusobola okugonjoola x$^2$ + 2 = 6, tusooka okuggyako 2 ku buli luuyi, ne tufuna x$^2$ = 4. Okukola ku x, tukubisa 4, ne tufuna x = ±2. N'olwekyo, x eba 2 oba -2. \\ 
& (\emph{To solve x$^2$ + 2 = 6, we first subtract 2 from both sides, resulting in x$^2$ = 4. To solve for x, we take the square root of 4, resulting in x = ±2. Therefore, x is either 2 or -2.}) \\
\midrule
\bf Prompt: & Nsemeriire kukora ki omwana wange obu araaba aine omuswija? \\
 & (\emph{Rutooro: What should I do if my child has a fever?}) \vspace{0.1cm} \\
\bf Response: & Kirungi kumanya ngu omuswija n'oburwaire bw'amaani oburikubaasa kuba obw'akabi.
\vspace{.5em}

Aha hariho ebi orikubaasa kukora kuhwera omwana waawe kwehurira gye:
\vspace{.5em}

Okunywa amaizi: Omwana waawe yaaba aine omuswija gw'amaani, oshemereire kumuha amaizi maingi kugira ngu atagira ekyanda.
[...] \vspace{0.5em} \\
& (\emph{It is well to know that malaria is a serious disease that can be life-threatening.}
\vspace{.5em}

\emph{Here are some things you can do to help your child feel better}:
\vspace{.5em}

\emph{Drink water: If your child has a high fever, you should give them plenty of water to prevent them from getting dehydrated.} [...]) \\
\bottomrule
\end{tabular}
\label{tab:prompt_and_response_examples}
\end{table}

\subsection{Machine translation}
\label{sec:translation_eval}

The test set covers realistic communication scenarios in Ugandan contexts, comprising parallel sentences spanning diverse domains: healthcare consultations, banking transactions, educational content, agricultural advice, market interactions, emergency communications, and civil service procedures. This design reflects practical deployment scenarios.

We compare Sunflower against leading proprietary models: GPT-4o, Gemini 2.5 Pro, DeepSeek-Chat, and Grok-3. We also evaluated NLLB-1.3B \citep{costa2022no}, though NLLB's coverage includes only four of our target languages (Acholi, Luganda, Kinyarwanda and Swahili), limiting comprehensive comparison. All models were evaluated using identical prompts formatted for translation tasks.

We employ standard machine translation metrics for quantitative evaluation. Primary metrics include BLEU \citep{papineni2002bleu} and chrF \citep{popovic2015chrf}, with chrF providing evaluation for morphologically rich languages. Before computing these metrics we first apply basic text normalisation, lower-casing and stripping punctuation,

Table~\ref{tab:results_chrf_eng_to_xx} and Table~\ref{tab:results_chrf_xx_to_eng} present chrF scores for both translation directions across all 31 languages. Sunflower-32B achieves the highest mean chrF score (0.435) for local-to-English translation, substantially outperforming Gemini 2.5 Pro (0.408), GPT-4o (0.354), and other baselines. For English-to-local translation, Sunflower-14B achieves competitive performance (0.366 mean chrF) while requiring significantly fewer parameters than proprietary alternatives.

\begin{table}[t]
\footnotesize
\centering
\caption{Average translation performance across 31 Ugandan languages for leading models.}
\label{tab:avg_performance}
\begin{tabular}{@{}lcccc@{}}
\toprule
\multirow{2}{*}{\textbf{Model}} & \multicolumn{2}{c}{\textbf{xx $\rightarrow$ eng}} & \multicolumn{2}{c}{\textbf{eng $\rightarrow$ xx}} \\
\cmidrule(lr){2-3} \cmidrule(lr){4-5}
& chrF & BLEU & chrF & BLEU \\
\midrule
Sunflower-32B & \textbf{0.435} & \textbf{20.625} & 0.357 & 7.598 \\
Sunflower-14B & 0.419 & 19.613 & \textbf{0.366} & \textbf{7.871} \\
Gemini 2.5 Pro & 0.408 & 18.559 & 0.301 & 7.760\\
GPT-4o & 0.354 & 14.850 & 0.235 & 6.531 \\
Grok-3 & 0.347 & 13.508 & 0.247 & 6.315 \\
DeepSeek-Chat & 0.308 & 11.260 & 0.237 & 5.944 \\
\bottomrule
\end{tabular}
\label{tab:translation_metrics_leaders}
\end{table}

Figure~\ref{fig:comparison_chrf} shows the performance advantage: Sunflower-14B achieves highest accuracy in 24 of 31 languages when averaging bidirectional chrF scores. We observe consistent superiority across languages, even in different groups (Bantu, Nilotic, Central Sudanic).

\begin{table}[hb]
\footnotesize
\centering
\caption{Comparison of translation performance with open-source models. Metrics are averages across 31 languages in both xx $\rightarrow$ eng and eng $\rightarrow$ xx directions.}
\label{tab:translation_results}
\begin{tabular}{llcccc}
\toprule
Model & Params & BLEU & chrF & CER & WER \\
\midrule
\multicolumn{6}{l}{\textit{Large Models (20B+ parameters)}} \\
sunflower-32b & 32B & \textbf{14.38} & \textbf{0.400} & \textbf{0.579} & \textbf{0.834} \\
qwen-3-32b & 32B & 2.85 & 0.192 & 0.992 & 1.411 \\
gemma-3-27b & 27B & 1.79 & 0.227 & 3.199 & 3.730 \\
gpt-oss-20b & 20B & 0.64 & 0.161 & 5.109 & 6.224 \\
\addlinespace
\multicolumn{6}{l}{\textit{Medium Models (7-14B parameters)}} \\
sunflower-14b & 14B & \textbf{12.32} & \textbf{0.377} & \textbf{0.789} & \textbf{1.097} \\
gemma-2-9b & 9B & 2.46 & 0.218 & 1.656 & 2.323 \\
llama-3.1-8b & 8B & 1.97 & 0.187 & 2.433 & 3.063 \\
aya-expanse-8b & 8B & 1.32 & 0.191 & 1.725 & 2.139 \\
llama-3-8b & 8B & 1.09 & 0.182 & 2.514 & 3.293 \\
qwen-2.5-7b & 7B & 1.29 & 0.193 & 2.265 & 2.712 \\
\addlinespace
\multicolumn{6}{l}{\textit{Small Models (< 1B parameters)}} \\
InkubaLM-0.4B & 0.4B & 0.54 & 0.161 & 3.783 & 3.950 \\
\bottomrule
\end{tabular}
\begin{tablenotes}
\footnotesize
\item \textsuperscript{*}From Figure~\ref{fig:comparison_chrf}; evaluated on subset of languages.
\item Bold indicates best performance within size class. For CER and WER, lower is better.
\item Sunflower models substantially outperform all open-source models across all size classes.
\end{tablenotes}
\label{tab:translation_metrics_open_source}
\end{table}

\subsubsection{Knowledge and Reasoning}
\label{sec:knowledge_eval}

To evaluate capabilities beyond translation, we assessed Sunflower on AfriMMLU, a component of the Afrobench benchmark suite \citep{ojo2025afrobench}. AfriMMLU is a multilingual adaptation of the widely-used MMLU benchmark \cite{hendrycks2020measuring}, covering general knowledge and reasoning across diverse subjects including science, history, and mathematics. Of the 16 African languages included in AfriMMLU, three overlap with our target language set: Swahili (swa), Kinyarwanda (kin), and Luganda (lug). We report zero-shot performance on these languages as an indicator of the models' broader reasoning capabilities. Following the IrokoBench evaluation protocol, we evaluate models using the ``direct'' setting, where questions are presented in the African language without translation. All models are instruction-tuned variants to ensure fair comparison.

Table~\ref{tab:afrimmlu_irokobench} presents our results compared to recent open-source models of comparable size. Sunflower-32B achieves the highest overall performance with 41.7\% F1 Score averaged across the three languages, outperforming all baselines including the larger Gemma-2-9B (38.0\%) and Qwen3-32B (28.2\%). Notably, Sunflower-32B ranks first on both Luganda (37.0\%) and Kinyarwanda (38.6\%), demonstrating strong performance on the lower-resource languages that were primary targets of our training approach. Sunflower-14B achieves 35.6\% average F1 score, ranking third overall.

However on knowledge and reasoning, all open source models are substantially behind the proprietary model GPT 4o. As our focus in this work has been on translation performance, the Sunflower models are relatively unoptimised for this multiple choice task. Several methods such as alternative reinforcement learning techniques are available to try to close this gap, but we leave this to future work.

\begin{table*}[t]
\centering
\small
\caption{AfriMMLU accuracy results. Sunflower models perform well related to other open source models, but substantially behind the proprietary GPT 4o.}
\label{tab:afrimmlu_irokobench}
\begin{tabular}{lccccc}
\toprule
\textbf{Model} & \textbf{lug} & \textbf{kin} & \textbf{swa} & \textbf{ave} \\
\midrule
\multicolumn{5}{c}{\textit{Open source}} \\
Sunflower-32B & \bf 37.0 & \bf 38.6 & 49.4 & \bf 41.7 \\
Gemma-2-9b-it & 31.6 & 32.2 & \bf 50.2 & 38.0 \\
Sunflower-14B & 32.2 & 34.6 & 40.0 & 35.6 \\
Llama-3.1-8B-Instruct & 29.6 & 30.2 & 37.2 & 32.3 \\
Aya-Expanse (8B) & 29.6 & 28.2 & 30.2 & 29.3 \\
Qwen2.5-7B-Instruct & 28.4 & 28.2 & 31.0 & 29.2 \\
Qwen3-32B & 25.2 & 26.0 & 33.4 & 28.2 \\
Aya-Expanse-32b & 28.6 & 27.2 & 28.4 & 28.1 \\
Llama-3-8B-Instruct & 25.2 & 25.4 & 33.4 & 28.0 \\
InkubaLM-0.4B & 22.0 & 24.6 & 25.4 & 24.0 \\
Gpt-OSS-20b & 24.6 & 23.2 & 23.6 & 23.8 \\
Llama-2-7B & 24.4 & 22.6 & 21.6 & 22.9 \\
\midrule
\multicolumn{5}{c}{\textit{Closed source}} \\
GPT 4o$^*$ & \bf 52.8 & \bf 64.2 & \bf 77.4 & \bf 64.8 \\
\bottomrule
\end{tabular}
\begin{tablenotes}
\footnotesize
\item \textsuperscript{*}GPT 4o metrics reproduced from \cite{adelani2024irokobench}.
\end{tablenotes}
\end{table*}

\subsection{Community feedback}

\begin{figure}[ht]
    \centering
    \includegraphics[width=0.7\linewidth]{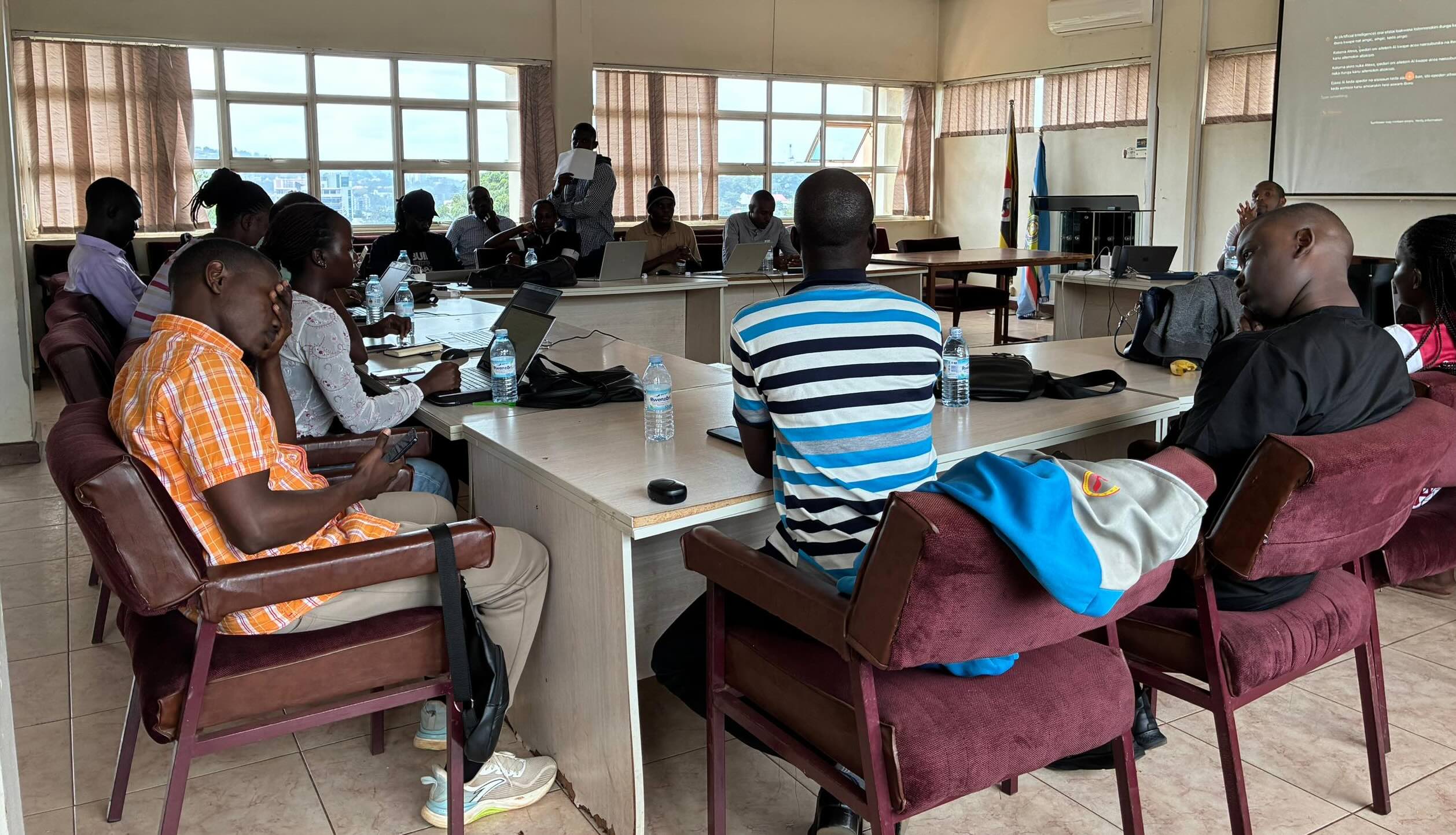}
    \caption{In-person testing of Sunflower by volunteers  speaking several different Ugandan languages.}
    \label{fig:in_person_testing}
\end{figure}

While our quantitative evaluation focuses on translation and multiple-choice question answering, Sunflower demonstrates broader multilingual capabilities, the models were tested by volunteer speakers of many languages in order to qualitatively assess them. Volunteers were involved in both online testing and in-person (Figure \ref{fig:in_person_testing}).

This testing focussed on abilities of the models as well as safety and bias. Issues reported by the testers were mitigated by creating additional training data, such as extra fine-tuning instruction examples. Particular limitations that we encountered:

\begin{itemize}
    \item Slang terms were generally not understood by the model.
    \item Although hallucinations were mitigated by reinforcement learning, they would still occur. The model would attempt to answer even if asked about fictional entities, and factual recall was mixed.
    \item The model would sometimes reply in the wrong language. This would especially be case if it was queried in a language with sparse training data, such as Dopadhola, in which case the model would fall back to a related language, Acholi, for which it had seen more training data.
    \item The translations were most accurate with at least one sentence of input. When users tried to translate individual words, there were often errors. If a word was meaningful in multiple languages, the model had no way to explain the alternatives and would give only one.
    \item Code could be generated, e.g. in Python, with variable names and explanation in Ugandan languages, though we observed that local-language comments placed in the code were sometimes unintelligible.
\end{itemize}

\begin{figure}[ht]
    \centering
    \includegraphics[width=0.4\linewidth]{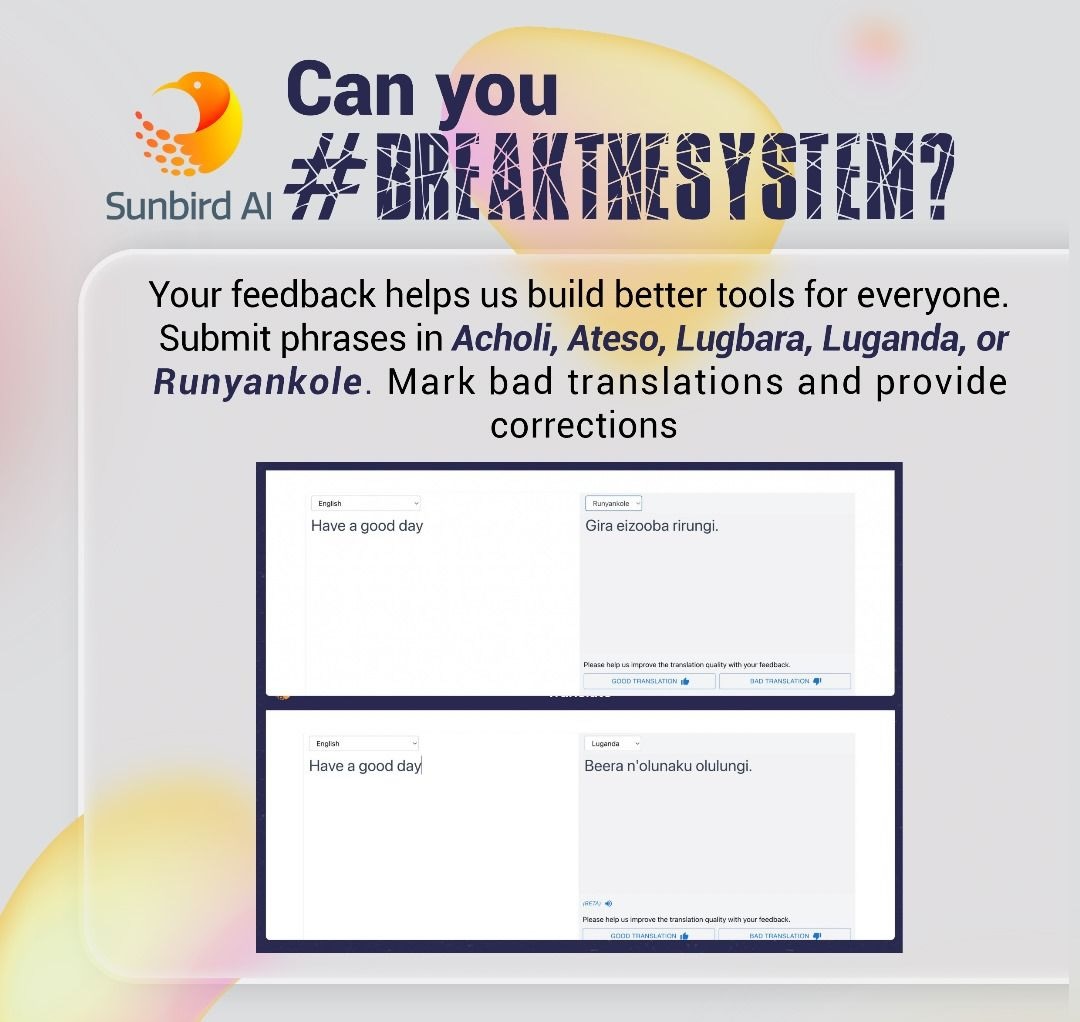}
    \hspace{0.5cm}
    \includegraphics[width=0.4\linewidth]{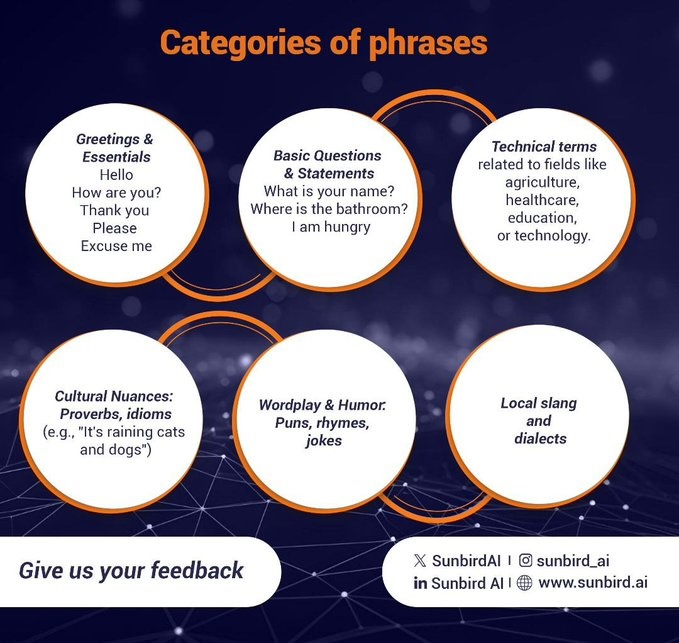}
    \caption{Feedback solicited from community members through the online \#breakthesystem campaign. This was particularly aimed at finding examples where translation performed poorly.}
    \label{fig:breakthesystem}
\end{figure}

\section{Discussion}
\label{sec:discussion}

In this paper, we have described the creation of the Sunflower 14B and 32B models, which have comprehension of nearly all Ugandan languages. Our focus is on machine translation, which is a useful measure of cross-lingual understanding and an important practical task, and we achieve state of the art translation accuracy for 24 of 31 languages spoken in Uganda. 

We also explain steps for instruction tuning and show that the model is capable of a wide variety of tasks, although the models are less optimised for this and future work on additional reinforcement learning will mitigate remaining issues such as hallucinations. Future work will also focus on grounding the model responses: for example, we have had encouraging initial results with incorporating web search results while generating answers.

In principle, the steps that we have outlined here could be carried out in any country, even those with very high linguistic diversity. We argue that this is a feasible path to LLM support for the majority of the world's languages: not as a centralised, monolithic architecture, but as a series of smaller, community-created models which meet the practical, communication and information needs of people in each region. 

\section*{Acknowledgments}

This work was made possible by funding from the Hewlett Foundation and IDRC Canada. We are grateful to our collaborators for feedback and assistance with datasets, including the Makerere University Centre for Language and Communication Services, Makerere AI Lab, Backup Uganda, Trac FM, SEMA Uganda, Simba FM and Cross-Cultural Foundation of Uganda.

\bibliographystyle{unsrt}
\bibliography{references}

\appendix
\section{Appendix A: Extended evaluation metrics}
Extended tables/plots.

\begin{table}[h!]
\footnotesize
\centering
\begin{tabular}{llcccccc}
\toprule
\multicolumn{2}{c}{Source language} & sunflower-32b & sunflower-14b & gemini-2.5-pro & gpt-4o & deepseek-chat & grok-3 \\
\midrule
ach & Acholi & \textbf{0.530} & 0.526 & 0.501 & 0.318 & 0.256 & 0.361 \\
adh & Jopadhola & \textbf{0.388} & 0.371 & 0.361 & 0.263 & 0.211 & 0.290 \\
alz & Alur & \textbf{0.423} & 0.381 & 0.333 & 0.260 & 0.232 & 0.275 \\
bfa & Bari & \textbf{0.358} & 0.343 & 0.259 & 0.254 & 0.249 & 0.254 \\
cgg & Rukiga & \textbf{0.525} & 0.487 & 0.524 & 0.466 & 0.411 & 0.420 \\
gwr & Lugwere & 0.480 & 0.474 & \textbf{0.490} & 0.448 & 0.344 & 0.408 \\
kdi & Kumam & \textbf{0.488} & 0.479 & 0.410 & 0.298 & 0.242 & 0.308 \\
kdj & Karamojong & \textbf{0.316} & 0.274 & 0.246 & 0.210 & 0.194 & 0.214 \\
keo & Kakwa & \textbf{0.301} & 0.284 & 0.233 & 0.233 & 0.230 & 0.239 \\
kin & Kinyarwanda & 0.572 & 0.577 & \textbf{0.592} & 0.580 & 0.540 & 0.555 \\
koo & Rukonjo & \textbf{0.366} & 0.322 & 0.355 & 0.291 & 0.255 & 0.264 \\
kpz & Kupsabiny & 0.210 & 0.194 & 0.222 & 0.197 & 0.211 & \textbf{0.222} \\
laj & Lango & \textbf{0.524} & 0.511 & 0.484 & 0.322 & 0.255 & 0.379 \\
lgg & Lugbara & \textbf{0.452} & 0.446 & 0.250 & 0.227 & 0.224 & 0.260 \\
lsm & Samia & 0.385 & 0.360 & \textbf{0.386} & 0.324 & 0.272 & 0.308 \\
luc & Aringa & 0.252 & \textbf{0.263} & 0.224 & 0.235 & 0.224 & 0.242 \\
lug & Luganda & 0.596 & 0.616 & \textbf{0.632} & 0.606 & 0.528 & 0.577 \\
mhi & Ma'di & \textbf{0.278} & 0.259 & 0.237 & 0.223 & 0.227 & 0.246 \\
myx & Lumasaba & \textbf{0.465} & 0.425 & 0.394 & 0.363 & 0.256 & 0.316 \\
nuj & Lunyole & \textbf{0.400} & 0.379 & 0.379 & 0.322 & 0.279 & 0.325 \\
nyn & Runyankole & 0.517 & 0.525 & \textbf{0.529} & 0.456 & 0.398 & 0.431 \\
nyo & Runyoro & 0.576 & 0.555 & \textbf{0.596} & 0.505 & 0.410 & 0.451 \\
pok & Pokot & 0.180 & 0.180 & \textbf{0.204} & 0.192 & 0.189 & 0.195 \\
rub & Lugungu & \textbf{0.504} & 0.467 & 0.456 & 0.355 & 0.308 & 0.344 \\
ruc & Ruruuli & 0.517 & 0.504 & \textbf{0.518} & 0.485 & 0.380 & 0.450 \\
rwm & Kwamba & 0.220 & 0.221 & \textbf{0.261} & 0.244 & 0.230 & 0.237 \\
swa & Swahili & 0.690 & 0.710 & \textbf{0.744} & 0.731 & 0.709 & 0.735 \\
teo & Ateso & \textbf{0.467} & 0.443 & 0.330 & 0.240 & 0.230 & 0.253 \\
tlj & Lubwisi & \textbf{0.391} & 0.359 & 0.383 & 0.355 & 0.291 & 0.302 \\
ttj & Rutooro & 0.530 & 0.493 & \textbf{0.533} & 0.442 & 0.373 & 0.404 \\
xog & Lusoga & \textbf{0.581} & 0.553 & 0.576 & 0.518 & 0.396 & 0.481 \\
\midrule
\multicolumn{2}{l}{Mean} & \textbf{0.435} & 0.419 & 0.408 & 0.354 & 0.308 & 0.347 \\
\bottomrule
\end{tabular}
\caption{chrF scores for translation directions xx $\rightarrow$ English.}
\label{tab:results_chrf_xx_to_eng}
\end{table}

\begin{table}[h!]
\footnotesize
\centering
\begin{tabular}{llcccccc}
\toprule
\multicolumn{2}{c}{Target language} & sunflower-32b & sunflower-14b & gemini-2.5-pro & gpt-4o & deepseek-chat & grok-3 \\
\midrule
ach & Acholi & 0.488 & \textbf{0.496} & 0.398 & 0.251 & 0.252 & 0.283 \\
adh & Jopadhola & 0.307 & \textbf{0.308} & 0.240 & 0.205 & 0.181 & 0.196 \\
alz & Alur & 0.267 & \textbf{0.285} & 0.206 & 0.172 & 0.171 & 0.179 \\
bfa & Bari & 0.221 & \textbf{0.277} & 0.166 & 0.066 & 0.110 & 0.152 \\
cgg & Rukiga & 0.445 & \textbf{0.450} & 0.399 & 0.288 & 0.270 & 0.268 \\
gwr & Lugwere & 0.360 & \textbf{0.370} & 0.326 & 0.302 & 0.291 & 0.279 \\
kdi & Kumam & 0.347 & \textbf{0.362} & 0.254 & 0.149 & 0.182 & 0.199 \\
kdj & Karamojong & \textbf{0.297} & 0.268 & 0.229 & 0.102 & 0.171 & 0.187 \\
keo & Kakwa & 0.264 & \textbf{0.315} & 0.169 & 0.092 & 0.146 & 0.160 \\
kin & Kinyarwanda & 0.546 & 0.538 & 0.568 & \textbf{0.581} & 0.448 & 0.481 \\
koo & Rukonjo & 0.344 & \textbf{0.355} & 0.258 & 0.232 & 0.224 & 0.224 \\
kpz & Kupsabiny & 0.182 & \textbf{0.184} & 0.177 & 0.078 & 0.143 & 0.147 \\
laj & Lango & 0.345 & \textbf{0.363} & 0.300 & 0.212 & 0.215 & 0.243 \\
lgg & Lugbara & 0.393 & \textbf{0.407} & 0.188 & 0.104 & 0.157 & 0.175 \\
lsm & Samia & 0.302 & \textbf{0.318} & 0.278 & 0.238 & 0.231 & 0.196 \\
luc & Aringa & \textbf{0.242} & 0.224 & 0.169 & 0.095 & 0.143 & 0.151 \\
lug & Luganda & 0.567 & 0.563 & \textbf{0.573} & 0.472 & 0.399 & 0.438 \\
mhi & Ma'di & 0.250 & \textbf{0.267} & 0.161 & 0.084 & 0.144 & 0.156 \\
myx & Lumasaba & \textbf{0.344} & 0.339 & 0.273 & 0.218 & 0.233 & 0.220 \\
nuj & Lunyole & 0.335 & \textbf{0.342} & 0.293 & 0.243 & 0.244 & 0.244 \\
nyn & Runyankole & 0.471 & \textbf{0.489} & 0.418 & 0.292 & 0.272 & 0.274 \\
nyo & Runyoro & 0.481 & \textbf{0.493} & 0.415 & 0.298 & 0.277 & 0.272 \\
pok & Pokot & \textbf{0.161} & 0.153 & 0.157 & 0.075 & 0.140 & 0.157 \\
rub & Lugungu & 0.325 & \textbf{0.366} & 0.216 & 0.208 & 0.195 & 0.197 \\
ruc & Ruruuli & \textbf{0.382} & 0.355 & 0.347 & 0.331 & 0.309 & 0.303 \\
rwm & Kwamba & 0.186 & \textbf{0.206} & 0.204 & 0.182 & 0.180 & 0.185 \\
swa & Swahili & 0.634 & 0.647 & 0.683 & \textbf{0.723} & 0.669 & 0.698 \\
teo & Ateso & 0.407 & \textbf{0.419} & 0.236 & 0.124 & 0.182 & 0.191 \\
tlj & Lubwisi & 0.306 & \textbf{0.322} & 0.275 & 0.239 & 0.224 & 0.228 \\
ttj & Rutooro & \textbf{0.436} & 0.435 & 0.367 & 0.285 & 0.248 & 0.256 \\
xog & Lusoga & 0.437 & \textbf{0.439} & 0.376 & 0.354 & 0.306 & 0.323 \\
\midrule
\multicolumn{2}{l}{Mean} & 0.357 & \textbf{0.366} & 0.301 & 0.235 & 0.237 & 0.247 \\
\bottomrule
\end{tabular}
\caption{chrF scores for translation directions English $\rightarrow$ xx.}
\label{tab:results_chrf_eng_to_xx}
\end{table}

\begin{table}[h!]
\footnotesize
\centering
\begin{tabular}{llcccccc}
\toprule
\multicolumn{2}{c}{Source language} & sunflower-32b & sunflower-14b & gemini-2.5-pro & gpt-4o & deepseek-chat & grok-3 \\
\midrule
ach & Acholi & 29.551 & \textbf{29.677} & 26.127 & 12.418 & 8.134 & 13.838 \\
adh & Jopadhola & \textbf{15.113} & 12.972 & 13.671 & 7.401 & 4.706 & 8.033 \\
alz & Alur & \textbf{17.878} & 13.978 & 9.938 & 6.555 & 5.248 & 6.387 \\
bfa & Bari & \textbf{12.822} & 12.678 & 6.658 & 6.609 & 5.639 & 5.595 \\
cgg & Rukiga & \textbf{27.007} & 22.836 & 24.157 & 19.955 & 15.600 & 16.779 \\
gwr & Lugwere & \textbf{23.290} & 23.252 & 22.398 & 20.148 & 12.246 & 16.110 \\
kdi & Kumam & \textbf{22.090} & 20.633 & 16.415 & 9.336 & 5.099 & 8.812 \\
kdj & Karamojong & \textbf{10.823} & 7.152 & 7.249 & 5.500 & 4.061 & 4.809 \\
keo & Kakwa & \textbf{9.804} & 9.166 & 6.675 & 7.573 & 6.420 & 6.359 \\
kin & Kinyarwanda & 32.406 & 33.807 & \textbf{35.788} & 33.118 & 30.054 & 30.335 \\
koo & Rukonjo & \textbf{15.391} & 12.301 & 13.588 & 10.277 & 7.006 & 7.165 \\
kpz & Kupsabiny & 4.454 & 3.495 & \textbf{5.158} & 4.322 & 4.668 & 4.603 \\
laj & Lango & 26.579 & \textbf{27.113} & 22.973 & 11.542 & 7.305 & 15.186 \\
lgg & Lugbara & \textbf{19.393} & 18.689 & 5.755 & 5.219 & 4.618 & 6.697 \\
lsm & Samia & 15.517 & 14.931 & \textbf{16.584} & 13.212 & 9.290 & 11.628 \\
luc & Aringa & 7.951 & \textbf{8.471} & 5.672 & 7.192 & 5.997 & 6.203 \\
lug & Luganda & 37.046 & 39.192 & \textbf{40.354} & 38.567 & 31.981 & 34.853 \\
mhi & Ma'di & \textbf{7.154} & 6.522 & 6.089 & 5.460 & 5.405 & 6.448 \\
myx & Lumasaba & \textbf{22.979} & 20.904 & 18.851 & 15.087 & 6.901 & 12.115 \\
nuj & Lunyole & \textbf{17.223} & 15.377 & 13.977 & 9.999 & 8.308 & 10.204 \\
nyn & Runyankole & 25.763 & \textbf{26.531} & 24.936 & 19.226 & 14.921 & 17.208 \\
nyo & Runyoro & 33.564 & 31.758 & \textbf{33.630} & 25.286 & 16.223 & 20.283 \\
pok & Pokot & 3.438 & 3.763 & \textbf{6.021} & 5.578 & 4.584 & 4.475 \\
rub & Lugungu & \textbf{23.868} & 21.446 & 18.703 & 11.812 & 8.746 & 11.253 \\
ruc & Ruruuli & 27.173 & \textbf{27.875} & 27.204 & 23.541 & 16.250 & 21.770 \\
rwm & Kwamba & 4.693 & 4.366 & \textbf{7.114} & 6.388 & 4.999 & 5.913 \\
swa & Swahili & 47.173 & 49.192 & \textbf{53.671} & 52.092 & 49.070 & 52.510 \\
teo & Ateso & \textbf{21.178} & 20.444 & 11.868 & 6.501 & 5.452 & 5.741 \\
tlj & Lubwisi & \textbf{17.727} & 15.121 & 16.732 & 14.656 & 9.718 & 9.599 \\
ttj & Rutooro & \textbf{25.597} & 23.006 & 24.821 & 17.709 & 13.078 & 14.584 \\
xog & Lusoga & \textbf{34.737} & 31.352 & 32.567 & 28.060 & 17.319 & 23.258 \\
\midrule
\multicolumn{2}{l}{Mean} & \textbf{20.625} & 19.613 & 18.559 & 14.850 & 11.260 & 13.508 \\
\bottomrule
\end{tabular}
\caption{BLEU scores for translation directions xx $\rightarrow$ English.}
\label{tab:results_bleu_xx_to_eng}
\end{table}

\begin{table*}[t]
\centering
\small
\caption{AfriMMLU results in direct (question is posed in Luganda, Kinyarwanda or Swahili) and translate-test (question is posed in a version translated to English) scenarios.}
\label{tab:afrimmlu_irokobench}
\begin{tabular}{lccccc}
\toprule
\textbf{Model} & \textbf{lug} & \textbf{kin} & \textbf{swa} & \textbf{ave} \\
\midrule
\multicolumn{5}{l}{\textit{Prompt LLMs in African Language}} \\
google--gemma-3-27b-it & 35.3 & 41.8 & 52.2 & 43.1 \\
Sunflower-32B & 36.9 & 38.5 & 49.5 & 41.7 \\
google--gemma-2-9b-it & 31.3 & 32.0 & 50.3 & 37.9 \\
Sunflower-14B & 32.3 & 34.1 & 40.0 & 35.4 \\
meta-llama--Llama-3.1-8B-Instruct & 29.5 & 28.9 & 37.0 & 31.8 \\
Qwen--Qwen2.5-7B-Instruct & 28.6 & 28.0 & 31.1 & 29.2 \\
Aya-101 (8B) & 29.5 & 27.9 & 30.1 & 29.2 \\
CohereLabs--aya-expanse-32b & 28.3 & 26.9 & 28.5 & 27.9 \\
Qwen--Qwen3-32B & 24.4 & 24.9 & 33.1 & 27.5 \\
meta-llama--Meta-Llama-3-8B-Instruct & 24.6 & 24.3 & 32.7 & 27.2 \\
openai--gpt-oss-20b & 24.3 & 23.0 & 23.6 & 23.6 \\
InkubaLM-0.4B & 20.4 & 23.3 & 24.2 & 22.7 \\
Llama-2-7B & 23.6 & 21.3 & 21.6 & 22.2 \\
\midrule
\multicolumn{5}{l}{\textit{Translate-Test (Eval. in English)}} \\
google--gemma-3-27b-it & 39.9 & 47.3 & 53.2 & 46.8 \\
google--gemma-2-9b-it & 35.3 & 39.6 & 47.8 & 40.9 \\
meta-llama--Llama-3.1-8B-Instruct & 37.0 & 41.2 & 43.9 & 40.7 \\
Sunflower-32B & 35.3 & 38.2 & 46.7 & 40.1 \\
Qwen--Qwen2.5-7B-Instruct & 35.7 & 38.7 & 45.8 & 40.0 \\
Sunflower-14B & 34.0 & 35.8 & 45.8 & 38.5 \\
meta-llama--Meta-Llama-3-8B-Instruct & 33.3 & 34.6 & 41.5 & 36.5 \\
CohereLabs--aya-expanse-32b & 30.2 & 38.2 & 39.6 & 36.0 \\
Aya-101 (8B) & 29.3 & 34.5 & 40.9 & 34.9 \\
Qwen--Qwen3-32B & 27.6 & 31.0 & 37.2 & 31.9 \\
Llama-2-7B & 26.0 & 27.0 & 33.9 & 29.0 \\
openai--gpt-oss-20b & 25.7 & 27.1 & 27.0 & 26.6 \\
InkubaLM-0.4B & 23.0 & 21.5 & 25.0 & 23.2 \\
\bottomrule
\end{tabular}
\end{table*}

\begin{table}[t]
\centering
\small
\caption{xx$\rightarrow$en  NLLB vs Sunflower (Chrf) }
\label{tab:xx2en_three_langs}
\begin{tabular}{lccccc}
\toprule
\textbf{Model} & \textbf{lug} & \textbf{kin} & \textbf{swa} & \textbf{Avg} \\
\midrule
Sunflower-32B & \textbf{0.567 } & \textbf{0.572} & \textbf{0.690} & \textbf{0.609} \\
Sunflower-14B & \textbf{0.563} & \textbf{0.577} & \textbf{0.710} & \textbf{0.616} \\
NLLB-3.3B     & 0.506 & 0.547 & 0.664 & 0.572  \\
NLLB-1.3B     & 0.500 & 0.532 & 0.672  & 0.568  \\
\bottomrule
\end{tabular}
\end{table}

\begin{table}[t]
\centering
\small
\caption{en$\rightarrow$xx NLLB vs Sunflower (Chrf)}
\label{tab:en2xx_three_langs}
\begin{tabular}{lccccc}
\toprule
\textbf{Model}  & \textbf{lug} & \textbf{kin} & \textbf{swa} & \textbf{Avg} \\
\midrule
Sunflower-32B & \textbf{0.567} & \textbf{0.546} & \textbf{0.634} & \textbf{0.582} \\
Sunflower-14B & \textbf{0.563} & \textbf{0.538 } & \textbf{0.647} & \textbf{0.583} \\
NLLB-3.3B     & 0.520  & 0.753  & 0.637 & 0.637  \\
NLLB-1.3B     & 0.477  & 0.622  & 0.628  & 0.576 \\
\bottomrule
\end{tabular}
\end{table}

\end{document}